\documentclass[letterpaper, 10 pt, conference]{ieeeconf}
\IEEEoverridecommandlockouts
\pdfoutput=1

\usepackage[T1]{fontenc}
\usepackage[utf8]{inputenc}
\usepackage[backend=biber,
            style=ieee,
            mincitenames=1,
            maxcitenames=2,
            maxbibnames=6,
            doi=false,
            isbn=false]{biblatex}
\usepackage[linesnumbered,
            ruled,
            vlined,
            noend]{algorithm2e}
\usepackage{amssymb,amsmath,amsfonts}
\usepackage[pdftex]{graphicx}
\usepackage{subfig}
\usepackage{verbatim}
\usepackage{hyperref}
\usepackage{tikz}

\hypersetup{
    unicode=true,
    pdfborder={0 0 0},
    breaklinks=true,
    pdfauthor={R. Connor Lawson, Linda Wills, Panagiotis Tsiotras},
    pdftitle={GPU Parallelization of Policy Iteration RRT\#},
    pdfkeywords={motion planning, RRT, PI-RRT\#, parallelization, GPU}
}
\makeatletter
\newcommand{\removelatexerror}{\let\@latex@error\@gobble}
\makeatother

\graphicspath{{figs/}}
\DeclareGraphicsExtensions{.pdf,.jpeg,.png}

\newcommand{\pirrt}{\mbox{PI-RRT\#}} 
\renewcommand{\G}{G}
\newcommand{\R}{\mathbb{R}}
\newcommand{\Rplus}{\R_+}
\newcommand{\X}{\mathcal{X}}

\newcommand{\Xfree}{\X_{free}}
\newcommand{\xinit}{x_{init}}
\newcommand{\xgoal}{x_{goal}}

\newcommand{\set}[1]{\{\,#1\,\}}

\newcommand{\T}{\pi} 

\newcommand{\union}[2]{ #1 \cup #2 }

\newcommand{\ghat}{\hat{g}}
\newcommand{\gT}{g_\T}

\newcommand{\dJ}{\Delta{}g}

\SetKwProg{proc}{Procedure}{}{}
\SetKwFunction{extend}{Extend}
\SetKwFunction{replan}{Replan}
\SetKwFunction{improve}{Improve}
\SetKwFunction{evaluate}{Evaluate}
\SetKwFunction{gpureplan}{GpuReplan}
\SetKwFunction{gpuimprove}{GpuImprove}
\SetKwFunction{gpuevaluate}{GpuEvaluate}
\SetKwFor{Loop}{loop}{}{end}
\SetKwFor{RepTimes}{repeat}{times}{end}
\SetKw{break}{break}

\newcommand{\email}[1]{\href{mailto:#1}{#1}}

\title{GPU Parallelization of Policy Iteration RRT\#}
\IEEEaftertitletext{\vspace{-0.5em}}

\author{%
    \authorblockN{R. Connor Lawson\authorrefmark{1}}
    \and
    \authorblockN{Linda Wills\authorrefmark{1}}
    \and
    \authorblockN{Panagiotis Tsiotras\authorrefmark{2}}%
\thanks{This work has been supported by the National Science
Foundation under Grant No. DGE-1650044. 
}%
\thanks{\authorrefmark{1}School of Electrical and Computer Engineering, Georgia
Institute of Technology, Atlanta, GA 30332. \email{rconnorlawson@gatech.edu},
\email{linda.wills@ece.gatech.edu}}%
\thanks{\authorrefmark{2}Daniel Guggenheim School of Aerospace Engineering,
Institute for Robotics and Intelligent Machines,
Georgia Institute of Technology, Atlanta, GA 30332.
\email{tsiotras@gatech.edu}}%
}

\newcommand\copyrighttext{%
  \footnotesize \textcopyright{} 2020 IEEE.  Personal use of this material is
  permitted.  Permission from IEEE must be obtained for all other uses, in any
  current or future media, including reprinting/republishing this material for
  advertising or promotional purposes, creating new collective works, for resale
  or redistribution to servers or lists, or reuse of any copyrighted component
  of this work in other works.}
\newcommand\copyrightnotice{%
\begin{tikzpicture}[remember picture,overlay]
\node[anchor=south,yshift=10pt] at (current page.south) {\fbox{\parbox{\dimexpr\textwidth-\fboxsep-\fboxrule\relax}{\copyrighttext}}};
\end{tikzpicture}%
}

\begin{document}

\maketitle
\copyrightnotice
\thispagestyle{empty}
\pagestyle{empty}

\begin{abstract} 

    Sampling-based planning has become a \emph{de facto} standard for complex
    robots given its superior ability to rapidly explore high-dimensional
    configuration spaces.  Most existing optimal sampling-based planning
    algorithms are sequential in nature and cannot take advantage of wide
    parallelism available on modern computer hardware.  Further, tight
    synchronization of exploration and exploitation phases in these algorithms
    limits sample throughput and planner performance.  Policy Iteration RRT\#
    (\pirrt{}) exposes fine-grained parallelism during the exploitation phase,
    but this parallelism has not yet been evaluated using a concrete
    implementation.  We first present a novel GPU implementation of \pirrt{}'s
    exploitation phase and discuss data structure considerations to maximize
    parallel performance.  Our implementation achieves 3--4$\times$ speedup over
    a serial \pirrt{} implementation for a 77.9\% decrease in overall planning
    time on average.  As a second contribution, we introduce the
    Batched-Extension RRT\# algorithm, which loosens the synchronization present
    in \pirrt{} to realize independent 12.97$\times$ and 12.54$\times$ speedups
    under serial and parallel exploitation, respectively.

\end{abstract}

\IEEEpeerreviewmaketitle

\section{Introduction}\label{introduction}

Motion planning is foundational to robotics, and the subject has received much
study over several decades of research. Sampling-based planning~%
\autocite{lavalle1998, lavalle2006, karaman2011a, elbanhawi2014, gammell2015a,
janson2015} has become the \emph{de facto} choice for complex robots given its
superior ability to rapidly explore high-dimensional configuration spaces.  Yet,
the problem is far from solved, and new techniques continue to be developed.

As path planning algorithms improve, their applications become more varied and
demanding. Recently, the so-called Task And Motion Planning (TAMP,
e.g.~\autocite{kaelbling2011}) seeks to marry geometric path planning and
semantic constraint satisfaction, requiring very fast geometric planning on
novel scenarios as a subroutine to more sophisticated algorithms. Other
extensions such as kinodynamic planning recast planning with differential
constraints as a geometric planning problem in a higher dimensional space via
projection onto the constraint manifold. As the cost of planning in general
grows exponentially with dimension, faster algorithms enable planning for more
realistic dynamical systems.

Since Karaman and Frazzoli's seminal paper in 2011~\autocite{karaman2011a},
focus on sampling-based planning has turned to finding not merely feasible
paths, but optimal ones. Planners seek shortest paths in some sense -- typically
Euclidean distance, either in configuration space or in some projection thereof.
Optimizing the solution can be seen as an ``exploitation'' phase utilizing
connectivity information gained during sample generation and graph extension,
the ``exploration'' phase~\autocite{rickert2014}.

Existing single-query sampling-based planning algorithms rely on tightly
interleaved exploration and exploitation, using the result of one phase to
inform the next in an iterative fashion. Many proofs of correctness and
optimality depend intimately on this ordering with precise synchronization. For
example, the optimal Rapidly Exploring Random Tree (RRT*) algorithm requires
sequential sampling and extension to achieve Voronoi-biased ``rapid''
exploration of the configuration space, but rewires the solution tree after each
new sample to achieve asymptotic optimality.

Synchronization of exploration and exploitation imposes a bound on parallelism
of planning algorithms. Probabilistic completeness and asymptotic optimality
hold only as the number of samples approaches infinity. Exploitation ``in the
loop'' requires more work per iteration, reducing sample throughput.  Excellent
prior work, such as \autocite{bialkowski2011, pan2012a}, has focused on
parallelizing and accelerating expensive subproblems of path planning -- most
commonly collision detection and nearest neighbor search -- but these approaches
sidestep the problem of synchronized exploitation, which ultimately limits the
available parallelism.

\textcite{arslan2016a} introduce new opportunities for parallel exploitation.
Their PI-RRT\# planner exposes fine-grained parallelism within shortest-path
search through a policy iteration method. However, this algorithm still retains
tight synchronization and contains only theoretical predictions of parallel
performance.

This work builds on \autocite{arslan2016a} to move towards desynchronizing
exploration and exploitation, and to evaluate parallel performance of \pirrt{}
with a concrete implementation. We present a novel GPU implementation of
PI-RRT\# and compare its performance to a serial implementation. Additionally,
we reduce the tight synchronization between exploitation and exploration in
PI-RRT\# by introducing batched graph extension and discuss future opportunities
for further asynchronous exploitation with respect to exploration.

Our parallel GPU implementation results in 3--4$\times$ speedup in exploitation
for large problems and an average decrease of up to 77.9\% in overall planning
time compared to a serial implementation. Batched graph extension, by contrast,
achieves up to 12.97$\times$ and 12.54$\times$ reduction in planning time with serial and
parallel implementations, respectively, when compared to more tightly synchronized
exploitation.

\autoref{related-work} discusses related work, including existing parallelized
planning algorithms. \autoref{overview-pirrt} defines relevant concepts and
notation of the PI-RRT\# algorithm. \autoref{gpu-algorithm} presents a novel GPU
implementation for exploitation in PI-RRT\#. \autoref{berrt} further modifies
the algorithm to support batched graph extension. \autoref{validation} evaluates
the proposed planners. \autoref{future-work} discusses future directions before
concluding in \autoref{conclusion}.

\section{Related work}\label{related-work}

Existing sampling-based algorithms are many and varied. They can generally be
divided into two classes: single-query and multi-query.  Single query algorithms
discover a tree of solutions rooted at the initial query point. Because they
evaluate connectivity of the free planning space at query time, these algorithms
are more adaptable to dynamic environments and very high dimensional spaces
where complete coverage is infeasible. The prototypical single-query algorithm
is the Rapidly Exploring Random Tree (RRT)~\autocite{lavalle1998}. Algorithms in
this group find a collision-free path from a single starting point in
configuration space to a goal region by building a directed tree of
collision-free edges. Notable recent algorithms in this category include: RRT*,
the optimal variant~\autocite{karaman2011a}; Fast Marching Tree
(FMT*)~\autocite{janson2015}, which explores a collection of samples in
best-first order according to a selection heuristic; Batch Informed Trees
(BIT*)~\autocite{gammell2015a}, a generalization of FMT* to incorporate informed
sampling; and RRT\#~\autocite{arslan2013}, which uses value iteration-based
exploitation to discover the best achievable paths through the current samples.
PI-RRT\#, as the name implies, is a modification of RRT\# that uses policy
iteration instead of value iteration as in the original RRT\#.

Multi-query algorithms, by contrast, precompute a roadmap of the configuration
space and connect pairs of start and goal query points to the map before solving
a simple graph search. The separation of roadmap construction and path query
allows very fast query-time execution, but requires a large memory footprint and
expensive up-front computation. These algorithms are best suited for static
environments with small configuration spaces, where thorough coverage by a
roadmap is feasible. Algorithms in this category descend from Probabilistic Road
Maps (PRM)~\autocite{kavraki1996a}.

There has been some previous work in parallelizing optimal single-query path
planning.  C-FOREST~\autocite{otte2013} replicates the path planning procedures
and data structures over multiple CPUs, broadcasting solution paths to other
processes as they are discovered.  P-RRT*~\autocite{ichnowski2012} uses multiple
threads running RRT* on a shared lock-free data structure.
\textcite{bialkowski2011} exposes parallelism by offloading collision detection
in RRT* to the GPU, but does not otherwise modify the algorithm.  All of these
approaches retain the underlying coupling of exploration and exploitation. As
such, they are complementary to the proposed approach.

Group Marching Tree (GMT*)~\autocite{ichter2017}, akin to both FMT* and the
\(\Delta\)-stepping shortest path algorithm, expands multiple nodes per
iteration in fixed-radius waves, similar to the proposed approach, but can only
produce near-optimal paths. Further, GMT* operates on a precomputed batch of
samples and neighbors, where the current work includes both sampling and
neighbor detection at runtime. This design choice retains opportunity for
sampling-time optimizations such as Informed Sampling~\autocite{gammell2014} and
lazily evaluates nearest neighbors only as necessary.

\section{Overview of PI-RRT\#}\label{overview-pirrt}

\pirrt{} is an extension of the Rapidly-exploring Random Tree algorithm (RRT)
which uses explicit policy iteration to maintain an optimal policy $\T : V \to
E$ and cost-to-come $\gT : V \to \Rplus$ for each vertex $v \in V$ during
construction of the free-space connectivity graph $G = (V, E)$.  Each vertex is
associated with a state $x \in \Xfree$ in the obstacle-free configuration space,
and the cost of an edge in $G$ is the cost to traverse between the states
associated with its endpoints, denoted $c : E \to \Rplus$.  An admissible
heuristic~$h : V \to \Rplus$ provides a lower bound on the cost-to-go from a
given vertex to the goal, and with the cost-to-come $\gT$ is used to track a
promising set $B \subseteq V$ of vertices that could potentially be on the
optimal path.

A high-level overview of \pirrt{} can be found in Algorithm~\ref{alg:pirrt-old}.
\pirrt{} performs exploitation in both the \extend and \replan procedures, but
\replan is the more expensive and crucial exploitation.  The \extend procedure
extends the graph by random sampling and extension in the same manner as RRT.
This is the exploration step.  Once a new vertex $v$ is added, the local minimum
cost-to-go $\gT(v)$ and policy $\T(v)$ is found by considering edges to each
existing vertex in a small neighborhood.  This small exploitation determines
whether the new vertex is promising, in which case it is added to the set $B$,
triggering a \replan before adding the next sample.  \replan performs policy
iteration to convergence in order to propagate updated path information from new
vertices and maintain optimal $\T$ and $\gT$ over the whole promising set.

\begin{figure}[!t]
    \removelatexerror
    \begin{algorithm}[H]
        \DontPrintSemicolon
        \label{alg:pirrt-old}
        \caption{PI-RRT\#~\autocite{arslan2016a}}
        $V \gets \{ \xinit, \xgoal \}$; $E \gets \emptyset$; $G \gets (V,E)$\;
        $B \gets \emptyset$\;
        \For{$k := 1$ to $N$}
        {
            $(G, B', \gT) \gets $ \extend{$G, B, \gT$}\;
            \If{$|B'| > |B|$}
            {
                $B \gets $ \replan{$G, B', \xinit, \xgoal$}\;
            }
        }
        $(V,E) \gets G; E' \gets \emptyset$\;
        \ForEach{$v \in V$}
        {
            $E' \gets \union{E}{\set{\T(v)}}$\;
        }
    \end{algorithm}
\end{figure}

As this work parallelizes the \replan procedure on the GPU, the serial version
is reproduced for reference in Algorithm \ref{alg:pirrt-replan}.  First, the
\improve procedure evaluates all neighbors of each promising vertex (denoted
$neigh(\G,v)$) and selects the locally optimal parent node, updating policy $\T$
accordingly.  This is the same local relaxation performed on the new vertex
during \extend.  Next, \evaluate rebuilds $B$ and computes updated $\gT$ for all
promising vertices via a truncated breadth-first traversal of the policy tree.
This process is repeated until convergence at tolerance $\epsilon$
(Line~\ref{pirrt-convergence}).  Additional detail on \pirrt{} can be found
in~\autocite{arslan2016a}.

\begin{figure}[!t]
    \removelatexerror
    \begin{algorithm}[H]
        \DontPrintSemicolon
        \label{alg:pirrt-replan}
        \caption{\pirrt{} Policy Iteration}
        \proc{\replan{$\G$, $B$, $\xinit$, $\xgoal$}}{
            \Loop{}{
                $\T, \dJ \gets $ \improve{$\G$, $c$, $B$, $\gT$}\;
                \If{ $\dJ < \epsilon$ }{ \label{pirrt-convergence}
                    \break\;
                }
                $\gT, B \gets $\evaluate{$\G$, $c$, $\T$, $\gT$, $h$}\;
            }
        }
        \proc{\improve{$\G$, $c$, $B$, $\gT$}}{
            \ForEach{$v \in B$}{
                $\ghat \gets \gT(v)$; $\dJ \gets 0$\;
                \ForEach{$n \in neigh(\G, v)$}{          \label{improve:inner-loop-start}
                    \If{$c(n, v) + \gT(n) < \ghat$}{    \label{improve-comparison}
                        $\ghat \gets c(n, v) + \gT(n)$\;
                        $\T(v) \gets n$\;
                    }
                }                                       \label{improve:inner-loop-end}
                $\dJ \gets \max{(\dJ, \gT(v) - \ghat)}$\;
            }
            \Return{$(\T, \dJ)$}\;
        }
        \proc{\evaluate{$\G$, $c$, $\T$, $\gT$, $h$}}{
            $Q \gets \set{\xinit}$; $B \gets \emptyset$\; \label{q-def}
            \While{$|Q| > 0$} {
                $v \gets $ pop($Q$)\;
                \ForEach{$n \in child(\T, v)$} {
                    $\gT(n) \gets c(v, n) + \gT(v)$\;     \label{evaluate:float1}
                    \If{$h(v) + \gT(v) < \gT(\xgoal)$} {  \label{evaluate:float2}
                        push($Q, n$)\;
                        $B \gets \union{B}{\set{n}}$\;
                    }
                }
            }
            \Return{$\gT, B$}
        }
    \end{algorithm}
\end{figure}

\section{GPU Implementation}\label{gpu-algorithm}

\pirrt{} exposes parallelism during exploitation that is not present in previous
approaches.  In particular, the Jacobi relaxation in the \improve procedure is
directly parallelized by processing each vertex in parallel.  Additionally, the
\evaluate procedure is a truncated breadth-first tree traversal, in which each
vertex on the frontier can be evaluated in parallel.  This section presents a
novel GPU implementation of \replan to take advantage of this parallelism.  Note
that this parallel implementation does not change the algorithm, rather it aims
to make maximum use of parallelism already available.

\subsection{Data Movement and Data Structure Preparation}

As with any GPU algorithm, data structure and memory movement is of primary
importance, as low bandwidth data movement between the CPU and GPU is the
strictest memory bottleneck.~\autocite{nvidia2019} The data in this algorithm
can be represented as a graph $G = (V, E)$ with properties on its edges and
vertices.  The data structures chosen to store these properties must facilitate
both efficient bus utilization and convenient memory access patterns.  Data such
as the world description and the parameters of the cost function $c$ are
constant and therefore negligible.  Each vertex has the following properties:

\begin{enumerate} 
    \item Cost-to-come $g(v) \in \Rplus$
    \item Heuristic cost-to-go $h(v) \in \Rplus$ 
    \item Parent $p \in \{V \cup \emptyset\}$, s.t. $\T(v) = (p, v)$
    \item Promising $b \in \{0, 1\}$, s.t. $b = 1 \iff v \in B$
\end{enumerate} 
All vertex properties are stored in structure of arrays format.  Several of
these data ($g, p,$ and $b$) are transferred to and from and GPU at each
\replan, so performant transfer is highest priority.  Structure of arrays format
stores each property as a contiguous buffer, facilitating constant time access
and high bandwidth transfer.

The edges have only a single property: the cost of traversal~$c$.  This cost is
evaluated once for each edge during \extend in order to evaluate the locally
optimal $\gT$ and $\T$, and its value is cached for use during \replan.  This
removes the need to copy potentially high-dimensional state data to the GPU,
further reducing data transfer overhead.

Storage for the edges and edge properties requires careful consideration, as the
graph structure chosen will impact both the communication and computation time
of the algorithm.  Three graph structures were considered:

\subsubsection{Adjacency List Graph}

This structure has favorable memory access patterns for vertex expansion and
graph extension, as costs and outgoing edges for each vertex are stored in
dynamic arrays which provide (amortized) constant-time access.  This is the
format used for graph representation in the CPU benchmark implementation (See
\autoref{validation}).  However, synchronizing adjacency lists between the CPU
and GPU is prohibitively expensive.  While edge lists for each vertex are
contiguous in memory, the collection of lists is not, as each list must be able
to grow dynamically.  As a result, synchronization requires many small memory
transfers, incurring a significant bandwidth penalty.

\subsubsection{Coordinate List Graph} 

This structure requires only a few large transfers to synchronize, leading to
high bandwidth utilization, but incurs some overhead due to storing the source
node of each edge redundantly.  Compared to adjacency lists, vertex expansion in
this format is much more expensive, as edges are not necessarily contiguous per
vertex.  Without sorting or other additional structure, vertex expansion
requires searching the entire structure for edges with the desired source node.

\subsubsection{Compressed Sparse Row (CSR) Graph}

This structure has the advantage of contiguous edges per vertex, as in the
Adjacency List graph, and does not suffer the space penalty of Coordinate lists.
However, edges are not easily added to existing vertices in CSR. In general,
adding new edges requires a rebuild step that is linear in the size of the
graph.

Consideration of data flow yields helpful insight into data structure selection.
There exists a producer-consumer relationship between the \extend and \replan
procedures for edges and costs.  New edges and costs are generated but unused in
\extend, and received but unmodified in \replan.

Given the one-way dataflow, the best possible scenario for graph data is to
transfer each edge to the GPU exactly once.  Therefore, the CPU-side data
structure can be optimized strictly for write performance and fast memory
transfer.  We found that the Coordinate List format on the CPU best supports
this use case, as new edges can be appended unordered to the list.
Additionally, since any new vertex will have at most one edge to each neighbor,
Coordinate List format is more space efficient than CSR in this case.

For fast vertex expansion during the \improve kernel, data is converted to CSR
format after transfer to the GPU.  The new edges are sorted in linear time using
radix sort, and a parallel merge over the old and new edge-lists rebuilds the
CSR.  This rebuild occurs once per \replan, so the cost is amortized over many
vertex expansions.

\subsection{Policy Improvement Kernel}

During policy improvement, each promising vertex searches over its outgoing
edges for the least cost path using cost-to-go estimates for its neighbors based
on the current policy.  In this kernel, each CUDA thread is assigned a single
vertex in the promising set, as suggested in the original \pirrt{}
paper~\autocite{arslan2016a}.  Each thread searches all outgoing edges from its
assigned vertex for the lowest cost parent.  This option suffers from load
imbalance since each vertex may have a variable number of neighbors, causing
threads to terminate at different times.  A load-balanced implementation based
on the reduce-by-key primitive from the Thrust library~\autocite{bell2017},
however, resulted in slower execution than the unbalanced version in practice.

\subsection{Policy Evaluation Kernel}

\improve may change the parent of each promising vertex based on its local
neighborhood, but it does not propagate cost reductions caused by upstream
rewiring to the leaves of the policy tree.  To correct this inconsistency,
\evaluate walks the policy tree outward in a breadth-first fashion, computing a
globally consistent cost-to-come for every visited vertex.  Along the way, the
promising set is rebuilt using the newly updated cost-to-come, and non-promising
branches are pruned from the traversal.  Only promising vertices and their
neighbors are re-evaluated, since only these vertices are relevant to finding an
optimal path.

Many GPU implementations of Breadth First Search already exist.
State-of-the-art algorithms, such as~\autocite{wang2017}, use a single GPU
kernel launch to process a whole frontier of vertices in parallel.  Our
implementation follows this approach, but takes advantage of the fact that $\T$
is a simple tree, rather than a general graph.  Because of this added structure,
our implementation does not require notions of open and closed sets typical of
BFS, reducing GPU global memory pressure and increasing performance.
Additionally, each outgoing edge from any frontier will always lead to a unique
vertex.

\section{Batch-Extension PI-RRT\#}\label{berrt}

For our second contribution, we offer a key insight, namely, that tightly coupled 
exploration and exploitation leads to decreased sample throughput and therefore 
decreased planning performance.
This is due to the high cost of tightly synchronized exploitation in the loop.
In the example of \pirrt{}, policy iteration may require, in the worst case, a
number of iterations two greater than the length of the optimal path in edges
\autocite[Theorem 2]{arslan2016a}. 

To address this issue, we introduce Batched-Extension \pirrt{}~(BERRT\#,
Algorithm~\ref{alg:berrt}). This algorithm utilizes a batch size parameter, $S$,
which delays the expensive \replan operation until $S$ samples have been added
th the graph (Lines~\ref{alg:berrt-rep}--\ref{alg:berrt-rep-end}). Allowing the
user to select a batch size exposes a tunable balance between exploration and
exploitation. In the case $S = 1$, BERRT\# reduces to \pirrt{}. When $S = N$,
BERRT\# is a variant of PRM*, computing all samples and neighbors first followed
by a single shortest path query.

\subsubsection*{Correctness}

\textcite{arslan2016a}, in their proof that \replan results in an asymptotically optimal
policy over the set of promising vertices, consider $\G$ as a whole and an arbitrary initial 
policy. Thus the same theorem implies correctness of BERRT\#, 
which differs from \pirrt{} only in the number of samples per iteration and thus in 
the quality of its initial policies.

Indeed, because BERRT\# does not immediately propagate path improvements 
discovered during \extend to the remainder of the promising set, it is possible 
that within the \extend loop policy $\T$ and cost-to-come $\gT$ are temporarily 
suboptimal. However, ensuring that \replan runs on the complete graph after all desired
samples have been added recovers a fully optimal solution.

\begin{figure}[!t]
    \removelatexerror
    \begin{algorithm}[H]
        \DontPrintSemicolon
        \label{alg:berrt}
        \caption{Batched-Extension \pirrt{}}
        $V \gets \{ \xinit, \xgoal \}$\;
        $E \gets \emptyset$\;
        $G \gets (V,E)$\;
        $B \gets \emptyset$\;
        \For{$k := 1$ to $N / S$}
        {
            $B' \gets B$\;
            \RepTimes{$S$} { \label{alg:berrt-rep}
                $(G, B') \gets $ \extend{$G, B', \xinit, \xgoal$}\;
            } \label{alg:berrt-rep-end}
            \If{$|B'| > |B|$}
            {
                $B \gets $ \replan{$G, B', \xinit, \xgoal$}\;
            }
        }
        $(V,E) \gets G; E' \gets \emptyset$\;
        \ForEach{$v \in V$}
        {
            $E' \gets \union{E}{\set{p_v}}$\;
        }
    \end{algorithm}
\end{figure}

\begin{figure*}[!t]
    \centering
    \subfloat[Performance]{\includegraphics[width=0.5\linewidth]{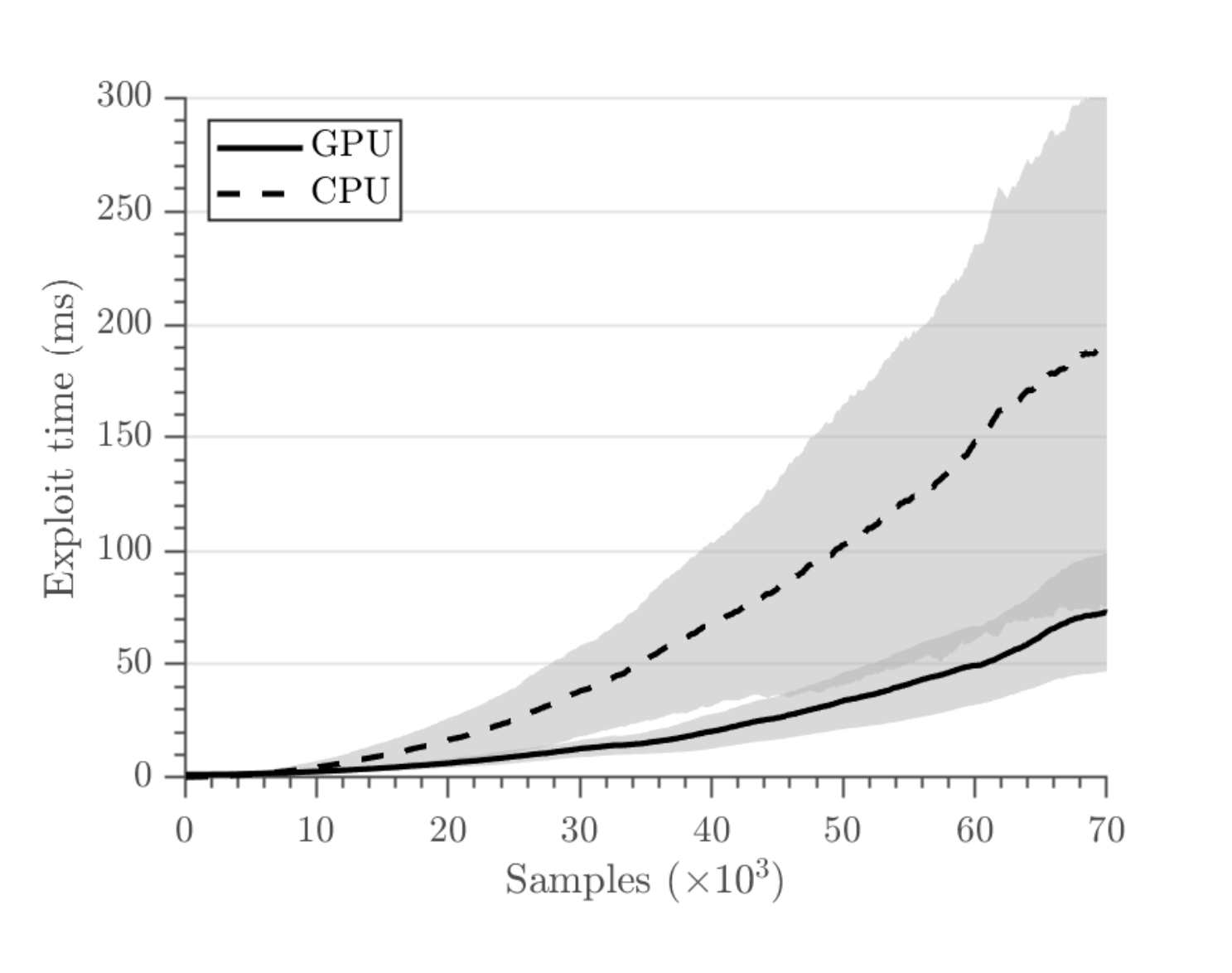}%
    \label{fig:break-even-A}}
    \hfil
    \subfloat[Speedup]{\includegraphics[width=0.5\linewidth]{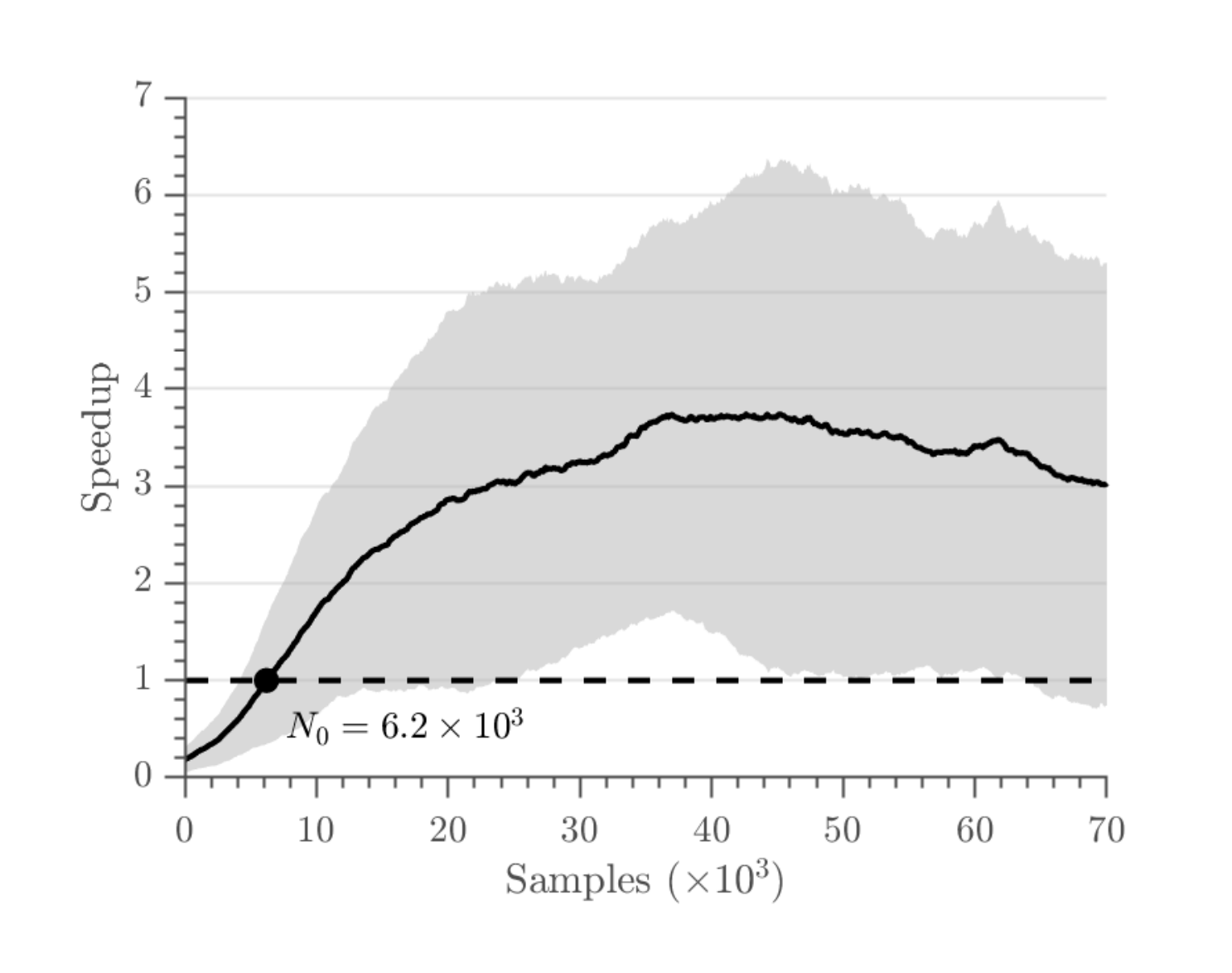}%
    \label{fig:break-even-B}}
    \caption{Performance comparison of parallel (GPU) and serial (CPU)
    exploitation. GPU exploitation outperforms CPU implementation above $N_0$,
    marked with a star.}
    \label{fig:break-even}
\end{figure*}

\section{Numerical Evaluation}\label{validation}

In this section, we analyze the performance of our GPU-based \replan against a
baseline CPU implementation using an Adjacency List graph. Further, we
investigate the impact of synchronized exploitation on planner performance by
varying the batch size in BERRT\# for cases of both serial and parallel
exploitation. Experiments were performed on a Intel Core i7-6820HQ 2.7 GHz CPU
running Ubuntu 16.04. GPU results were realized on an NVIDIA Quadro M2000M
mobile GPU with CUDA v10.2. Trials used independent random sample sequences on a
query for a two-dimensional system with polygonal obstacles. Each test was run
for five trials, and averaged results are reported. Where confidence intervals
are shown, they reflect one standard deviation from the reported mean.

First, we compare the performance of parallel exploitation to serial
exploitation. Figure~\ref{fig:break-even-A} shows the time spent for
exploitation as problem size increases for both implementations.
Given that the GPU implementation incurs overhead due to memory movement and
data structure reconstruction as discussed in \autoref{gpu-algorithm}, it is
unsurprising that the CPU-based implementation is faster on small problems.
Above the labelled threshold $N_0$, the superior performance of the GPU
algorithm is able to overcome this overhead. Figure~\ref{fig:break-even-B}
reports the relative speedup of the GPU exploitation per \replan. The peak
mean speedup is 3.74$\times$ at $N = 42,600$, before tapering back to 3$\times$ as
the problem size increases. More study is needed to understand this reduction in
relative performance on very large problems, but it can likely be attributed to the growing cost of
data movement and an increasingly random vertex access pattern during \evaluate,
reducing GPU utilization.

\begin{figure*}[!t]
    \centering
    \includegraphics[width=\linewidth]{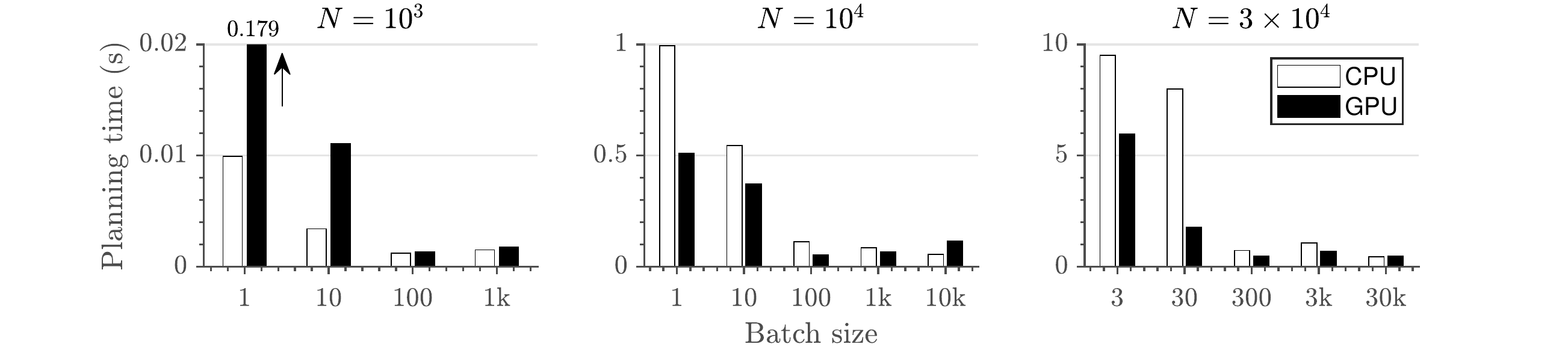}%
    \caption{Performance comparison of CPU and GPU exploitation for PI-RRT\#.
    Graphs are labelled with corresponding problem size, $N$. Increasing the batch size results in significant reduction in overall planning time at all problem sizes.}
    \label{fig:all_comparisons}
\end{figure*}

Figure~\ref{fig:all_comparisons} shows the average planning time for BERRT\#
under both serial and parallel exploitation implementations, across a variety of
problem sizes $N$ and batch sizes $S$. The problem size is the number of total 
samples considered before terminating the planner.

The number of samples in a batch has a profound effect on the performance of both 
the serial and parallel implementations, with larger batch sizes resulting in faster 
planner execution in most cases. When $N=10^4$, increasing the batch size one order of magnitude
from $S = 10$ to $S = 100$ results in a speedup of 4.84$\times$ on the CPU and
6.94$\times$ on the GPU. Compared to \pirrt{} ($S = 1$), BERRT\# with $S = 100$ achieves
an 8.83$\times$ speedup on CPU and 9.52$\times$ on GPU.

The largest observed speedup in this experiment was when $N=3\times 10^4$: 
increasing $S=3$ to $S=300$ resulted in
12.97$\times$ and 12.54$\times$ speedup on CPU and GPU implementations, respectively.
The largest decrease in total planning time due to only GPU
exploitation occurred when $N = 3 \times 10^4, S = 30$, a decrease of $77.9\%$.

The steep increase in performance with increasing batch size confirms the hypothesis
that tightly coupled exploitation results in low planner performance. However, there
is a limit to how much decoupling is effective. When $N=10^4$ and $N=3\times10^4$,
increasing the sample size from 1\% to 10\% of the total problem size resulted
in slight performance regressions. A possible cause of this regression is that larger
batches without replanning introduce more suboptimality to the policy, requiring more 
policy iterations to converge back to the optimal policy. Additionally, for more challenging 
search problems, sampling-time optimizations based on the results of incremental exploitation are more
important than in the 2D problem considered here. More frequent exploitation
allows new information to be incorporated into sampling more quickly.

\section{Future work}\label{future-work}

As noted above, due to data movement and data structure preparation overhead,
the proposed GPU-based implementation is only beneficial above a certain
threshold. This threshold likely depends not only on machine parameters such as
bus speed, memory bandwidth, and CPU cache size, but also on
difficult-to-observe problem parameters such as configuration-space
connectivity. Future work may investigate methods to adaptively select serial or
parallel exploitation in order to maximize performance across all regimes.

Another promising avenue is asynchronous policy iteration exploitation
concurrent with exploration. The BERRT\# algorithm loosens the synchronization
and demonstrates the potential speedup available, but the \replan procedure
still does not run concurrently with graph extension. Since our GPU
implementation successfully migrates policy iteration to a separate device, it
may proceed concurrently and asynchronously with graph extension. Updated policy
information and new edges in the graph may be exchanged periodically between
devices. This method would allow progress on both exploration and exploitation
to continue in parallel, and exposes opportunity to use complementary techniques
such as C-FOREST on the CPU to parallelize graph extension.

\section{Conclusion}\label{conclusion}

We have presented a GPU-based implementation of policy iteration for use in
sampling-based planning problems. The presented solution achieves an empirical
speedup of 3--4$\times$ for each iteration when compared to a serial implementation,
resulting in up to $77.9\%$ decrease in end-to-end planning time. Additionally,
employing the key insight that tightly synchronized exploitation slows sample
throughput, we present the Batched-Extension RRT\# algorithm and demonstrate its
effectiveness, reducing total planning time by up to an order of magnitude.

\printbibliography

\end{document}